\documentclass[10pt,twocolumn,letterpaper]{article}

\usepackage{cvpr}
\usepackage{times}
\usepackage{epsfig}
\usepackage{graphicx}
\usepackage{amsmath}
\usepackage{amssymb}
\usepackage{arydshln}
\usepackage{subfiles}
\usepackage{pdfpages}

\usepackage[pagebackref=true,breaklinks=true,letterpaper=true,colorlinks,bookmarks=false]{hyperref}

\cvprfinalcopy 

\def\cvprPaperID{4988} 

\ifcvprfinal\fi
\begin{document}

\title{Neural Point Cloud Rendering via Multi-Plane Projection}

\author{Peng Dai$^{1}$\thanks{Equal contribution}
\and Yinda Zhang$^{2*}$ 
\and Zhuwen Li$^{3*}$
\and Shuaicheng Liu$^{1}$\thanks{Corresponding author}
\and Bing Zeng$^1$\\
\and $^1$University of Electronic Science and Technology of China\\
$^2$Google Research \hspace{1cm} $^3$Nuro Inc\\
}


\maketitle
\thispagestyle{empty}

\begin{abstract}
   We present a new deep point cloud rendering pipeline through multi-plane projections. The input to the network is the raw point cloud of a scene and the output are image or image sequences from a novel view or along a novel camera trajectory. Unlike previous approaches that directly project features from 3D points onto 2D image domain, we propose to project these features into a layered volume of camera frustum. In this way, the visibility of 3D points can be automatically learnt by the network, such that ghosting effects due to false visibility check as well as occlusions caused by noise interferences are both avoided successfully. Next, the 3D feature volume is fed into a 3D CNN to produce multiple planes of images w.r.t. the space division in the depth directions. The multi-plane images are then blended based on learned weights to produce the final rendering results. Experiments show that our network produces more stable renderings compared to previous methods, especially near the object boundaries. Moreover, our pipeline is robust to noisy and relatively sparse point cloud for a variety of challenging scenes.
\end{abstract}

\section{Introduction}
Rendering is on high demand for many graphics and vision applications. To produce high quality rendering, various physical understanding of the scene has to be established, such as scene geometry~\cite{dai2017bundlefusion}, scene textures~\cite{blinn1978simulation}, materials~\cite{wang2009material}, illuminations~\cite{wood2000surface}, all of which require tremendous efforts to obtain. After the construction of rendering essentials, a photo-realistic view of the modeled scene is generated through expensive rendering process such as ray tracing~\cite{glassner1989introduction} and radiance
simulation~\cite{ward1994radiance}.

\begin{figure}[!htb]
\centering
\includegraphics[width=1.0\linewidth]{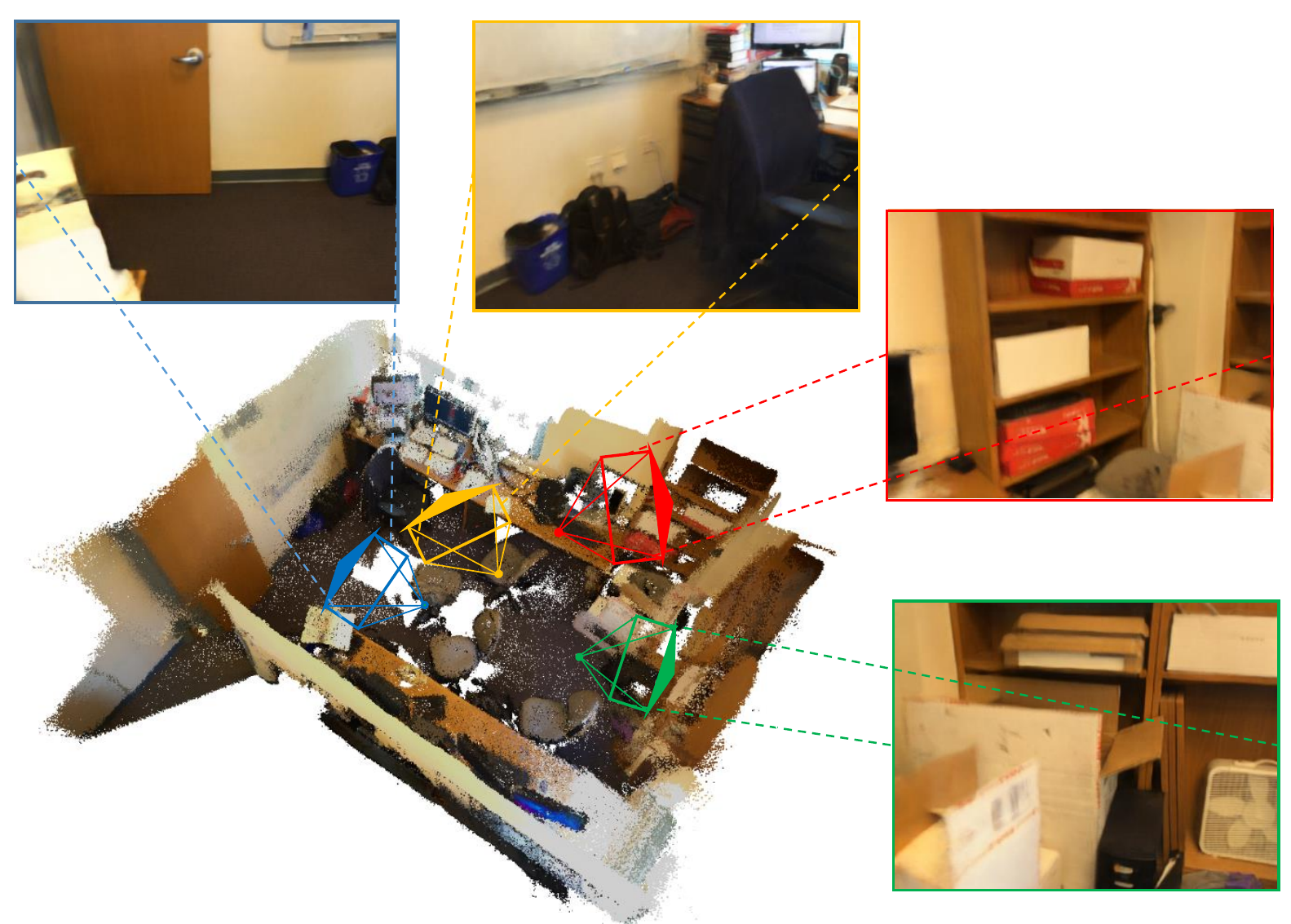}
\caption{Our method synthesize images in novel view by using neural point cloud rendering.}
\label{fig:teaser}
\vspace{-0.3cm}
\end{figure}
Image based rendering (IBR) techniques~\cite{gortler1996lumigraph,levoy1996light,mcmillan1995plenoptic}, alternatively, try to render a novel view based on the given images and their approximated scene geometries through image warping~\cite{liu2009content} and image inpainting~\cite{iizuka2017globally}. The scene structure approximation often adopts simplified forms such that the rendering process becomes relatively cheaper than physically based renderings. However, IBR requires the novel view to stay close to the original views in order to avoid rendering artifacts. Instead of purely based on images, point-based graphics(PBG) ~\cite{kobbelt2004survey,grossman1998point,gross2011point} simplify the scene structures by replacing the surface mesh with point cloud or surfels~\cite{pfister2000surfels}, such that the heavy geometry constructions can be avoid. 

On the other hand, many deep learning approaches show strong capability in inpainting~\cite{liu2018image}, refining~\cite{qiu2019deeplidar}, and even constructing images from only a few indications~\cite{park2019semantic,bau2019semantic}. Such capabilities can be considered as a complement to IBR, namely neural IBR~\cite{hedman2018deep}, to overcome the challenging during image synthesize. For examples, the empty regions caused by view change can be compensated through high quality inpainting~\cite{yu2019free}. 

Recently, combining advantages of simplified geometry representation and neural capabilities become a new trend,
yielding neural rendering methods~\cite{bui2018point,chen2018deep,hedman2018deep,hedman2018deep,nalbach2017deep}.
It directly learns to render end-to-end, which bypass complicated intermediate representations.
Previous work used mostly 3D volume as representation~\cite{sitzmann2019deepvoxels}.
However, the memory complexity of 3D volume is cubic, and thus these approaches are not scalable and usually only work for small objects. Recently, there is a trend to build neural rendering pipeline from 3D point cloud, which is more scalable for relatively larger scenes. However, 3D point cloud often contains strong noise due to both depth measurements~\cite{wolff2016point} and camera calibrations~\cite{zhang2014calibration}, which interference the visibility check when projected onto 2D image plane and results in jittering artifacts if a series of images are generated along a given 3D camera trajectory. On the other hand, this type of methods often require a lot of points for a reliable z-buffer check as well as the requirement of full coverage of all camera viewpoints. Even though the memory usage is linear with the number of points, the huge number still cause unaffordable memory and storage. Aliev~\emph{et al.}~\cite{aliev2019neural} proposed a neural point-based graphics approach that directly project 3D geometry onto the 2D plane for neural descriptor encoding, which not only ignores the visibility check but also suffers from noise interferences, resulting ghosting artifacts as well as strong temporal jitters.

In this paper, we propose a novel deep point cloud rendering pipeline through multi-plane projection, which is more robust to depth noise and can work with relatively sparse point cloud. In particular, instead of directly projecting features from 3D points onto 2D image domain using perspective geometry~\cite{hartley2003multiple}, we propose to project these features into a layered volume in the camera frustum. By doing so, all the features from points in the camera's field-of-view are maintained, thus useful features are not occluded accidentally by other points due to noisy interferences. The 3D feature volume are then fed into a 3D CNN to produce multiple planes of images, which corresponds to different space depths. The layered images are blended subsequently according to learned weights. In this way, the network can fix the point cloud errors in the 3D space rather than work on projected 2D images where visibility has already lost. In addition, the network can pick information accurately from the projected full feature volume to facilitate the rendering.

Extensive experiments evaluated on the popular dataset, such as ScanNet~\cite{dai2017scannet} and Matterport 3D~\cite{chang2017matterport3d}, show that our model produces more temporal coherent rendering results comparing to previous methods, especially near the object boundary. Moreover, the system can learn effectively from more points, but still performs reasonably well given relatively sparse point cloud.

To recap, we propose a deep learning based method to render images from point cloud. Our main contributions are summarized as:
\begin{itemize}
  \item 3D points are projected to a layered volume such that occlusions and noises can be handled appropriately.
  \item Not only the rendered single view is superior in terms of image quality, but also the rendered image sequences are temporally more stable.
  \item Our system works reasonably well with respect to relatively sparse point cloud.
\end{itemize}

\section{Related Work}

\subsection{Model-based rendering}
Model based rendering requires the construction of 3D models, such as multi-view structure-from-motion for point cloud recovery~\cite{hartley2003multiple}, surface reconstructiocn and meshing~\cite{wang2018pixel2mesh}.
When performance is preferred, ray tracing~\cite{glassner1989introduction} is used to simulate the transmission of light in the space, such that can better interact with the environment, e.g., geometry~\cite{snavely2006photo}, material~\cite{debevec1998efficient}, BRDF~\cite{goldman2009shape}, lighting~\cite{hold2017deep}, and produce more realistic rendering. However, each estimation step is prone to errors, which leads to the render artifacts. Moreover, such methods not only require a lot of knowledge of the scene, but also are notoriously slow.

\subsection{Image-based rendering}
Image based rendering(IBR)~\cite{gortler1996lumigraph,levoy1996light,mcmillan1995plenoptic,seitz1996view} aims to produce a novel view from given images through warping~\cite{liu2009content} and blending~\cite{debevec1998efficient}, which is a computationally efficient approach compared with classical rendering pipeline. Multiple view geometry~\cite{hartley2003multiple} is applied for camera parameter estimation or some variant that bypasses the 3D reconstruction~\cite{snavely2006photo}, such as adopting epipolar constraints~\cite{goldstein2012video}. Recently, deep learning has been proved to be more effective in replacing the warping and blending of traditional approaches in the IBR pipeline~\cite{flynn2016deepstereo,hedman2018deep}. However, the quality of rendered novel view still depends heavily on the distribution of existing views, sparse samples or large viewpoint drift would produce unsatisfactory results. Adopting light field cameras is one solution to alleviate such problems~\cite{kalantari2016learning}.

\subsection{Deep image synthesis}
Deep methods for 2D image synthesize has achieved very promising results, such as autoencoders~\cite{hinton2006reducing}, PixelCNN~\cite{van2016conditional} and image-to-image translation~\cite{isola2017image}. The most exciting results are produced based on generative adversarial networks~\cite{zhu2017unpaired,park2019semantic}. Most of the generators adopt encoder-decode architecture with skip connections to facilitate feature propagation~\cite{ronneberger2015u}. However, these approaches cannot be directly applied to the rendering task, as the underlying 3D structures cannot be exploited for 2D image translations.

\begin{figure*}[!htb]
\centering
\includegraphics[width=1.0\linewidth]{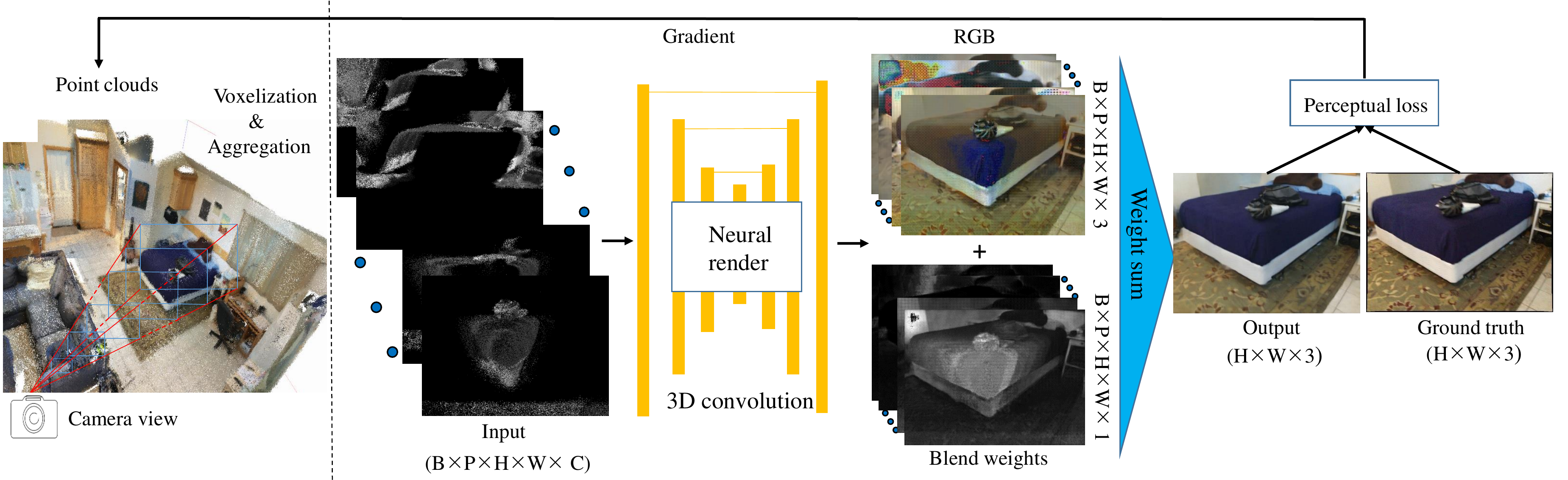}
\caption{Overview of our proposed method. Our method is divided into two parts, the multi-plane based voxelization (left) and multi-plane rendering(right). For the first part, point clouds are re-projected into camera coordinate system to form frustum region and voxelization plus aggregation operations are adopted to generate a multi-plane 3D representation, which will be concatenated with normalized view direction and sent to render network. For the second part, the concatenated input is feed into a 3D neural render network to predict the product with 4 channels (i.e. RGB + blend weight) and the final output is generated by blending all planes. The training process is under the supervision of perceptual loss, and both network parameters and point clouds features are optimized according to the gradient.}
\label{fig:network}
\vspace{-.5em}
\end{figure*}

\subsection{Neural rendering}
Recently, deep learning are used to renovating the rendering~\cite{thies2019deferred}.
Nvidia~\cite{chaitanya2017interactive} uses deep learning to denoise a relatively fast low-quality rendering.
More fundamentally, many successes have been achieved by neural rendering that directly learn representation from input and produce desired output, such as DeepVoxel~\cite{sitzmann2019deepvoxels} and Neural Point Based Graphics(NPG)~\cite{aliev2019neural}. Most of them rely on a 3D volume intermediate representation. The DeepVoxel cannot render large scenes, such as room environment. The most related work is NPG~\cite{aliev2019neural}; it also proposes to render images from point cloud, by projecting learned features from points to the 2D image plane according to perspective geometry, and train a 2D CNN to produce the color image.
The method learns complete and view-dependent appearance, however, it suffers from visibility verification problem due to directly 3D projection, and is also sensitive to point cloud noises. We project 3D points to a layered 3D volume to overcome such problems.

\section{Method}
\vspace{-0.2em}
\subsection{Overview}
Our deep learning framework receives a point cloud representation of a scene and generates photo-realistic images from an arbitrary camera viewpoint. The overview of the framework is illustrated in Fig.~\ref{fig:network}. The whole framework consists of two modules: multi-plane based voxelization and multi-plane rendering.
The multi-plane based voxelization module divides the 3D space of the camera view frustum into voxels w.r.t. image dimensions and a pre-defined number of depth planes. The voxels then aggregates features of points inside it with geometric rules.
The 3D feature volume is then fed into the multi-plane rendering module, which is a 3D CNN, to generates one color image plus a blending weight per plane on the depth dimension of the volume.
The final output is a weighted blending of multi-plane images.
It is worth noting that the point cloud feature representation and the network are jointly optimized in an end-to-end fashion.
The remaining part of this section describes details of these two modules. 

\subsection{Learnable point cloud features}
Our input is a 3D point cloud representation of the scene.
To be sufficient for rendering, each 3D point should contains both position and appearance feature. 
The position feature is obtained from 3D reconstruction, and one simple way to collect appearance feature is to keep the RGB value from the corresponding image pixel. However, one point may show different RGB intensities when observed from different views due to view-depend effects (\eg, reflection and highlight). To solve this problem, we learn a 8-dimensional vector as the appearance feature jointly with the network parameters.
To this end, we update this feature by propagating gradient to the input \cite{aliev2019neural, thies2019deferred}, such that the appearance feature can be automatically learned from data. 

Since the object appearance is often view-depend, we further take view direction between camera position and point clouds in 3D space into consideration. Thus, we concatenate the normalized view direction of the point as an additional feature vector to each point, following~\cite{aliev2019neural}. Note that this feature is not trainable as the point position.

\subsection{Multi-plane based voxelization}
\begin{figure}[!htb]
\centering
\includegraphics[width=1.0\linewidth]{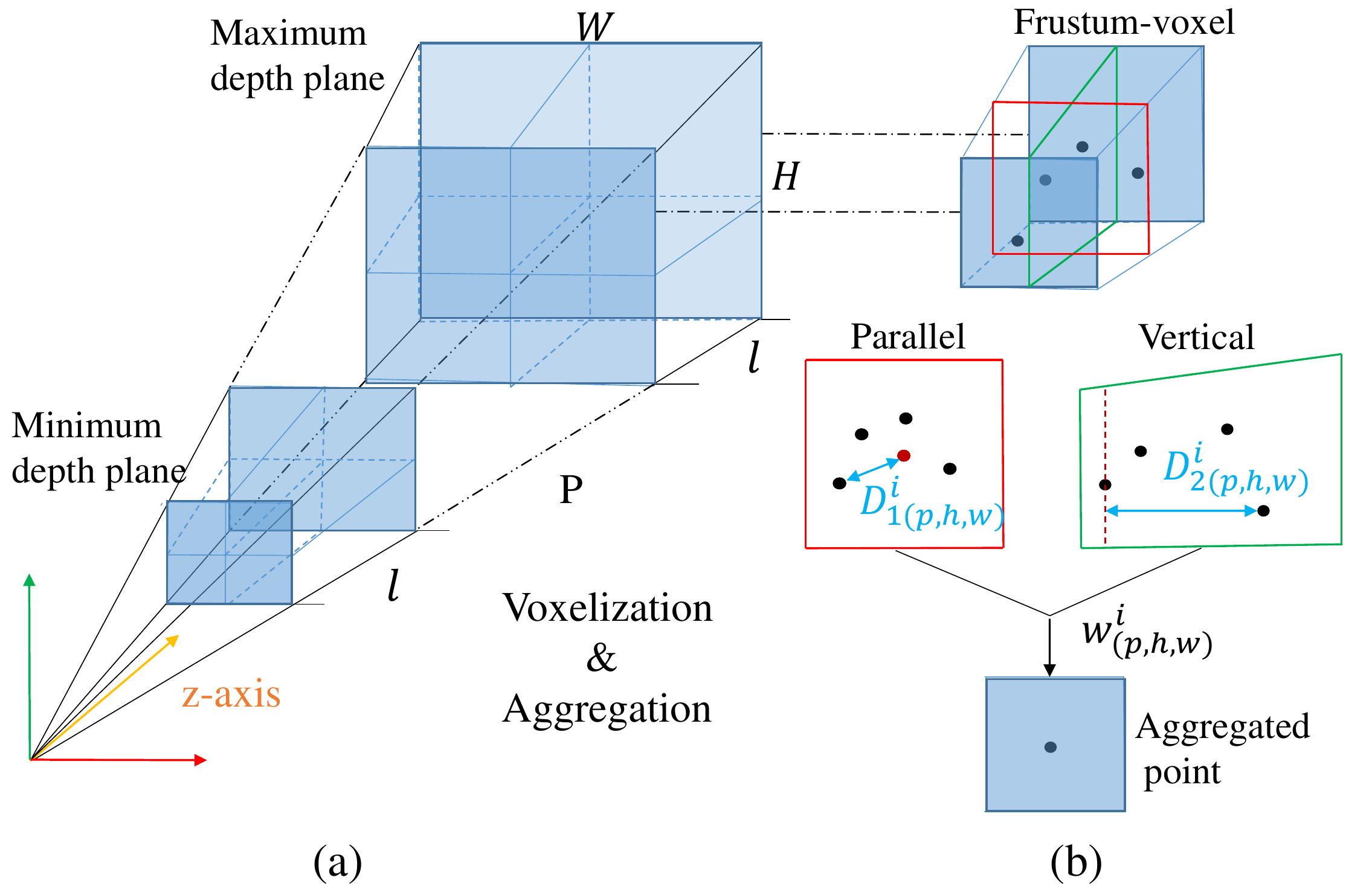}
\caption{3D space voxelization and aggregation. (a) Multi-plane voxelization. Point clouds under camera view are restricted into a frustum region specified by the minimum and maximum depth planes, and such region can be uniformly split into small frustum voxels acoording to image size$(H,W)$ and predefined number of planes $P$, where $l$ indicates length of frustum. (b) Aggregation. To illustrate, we take one frustum voxel for example. In oredr to aggregate points in one frustum voxel, we calculate two kinds of distance. $D^i_{1(p,h,w)}$ is the distance between points and pixel center in parallel direction and  $D^i_{2(p,h,w)}$ calculates depth difference between points and minimum depth value in this voxel in vertical direction. Further, we can define the blend weight $w^i_{(p,h,w)}$ of each point on the basis of these two distance. Finally, the blend weights are applied to aggregate multiple points into a new point. }
\label{fig:division and integartion}
\end{figure}

\noindent
\textbf{Layered voxels on camera frustum.} 
The image dimension is denoted as $H\times W$.
With a known camera projection matrix, each pixel is lifted to 3D space to form a frustum with near and far planes specified by the minimum and maximum depth of the projected point cloud. The frustum is further divided into $P$ small ones along the z-axis uniformly as shown in Fig.~\ref{fig:division and integartion} (a). As a result, we will obtain $P\times H\times W$ frustum voxels.

\noindent
\textbf{Feature aggregation} 
The next step is to aggregate feature from the point cloud into the camera frustum volume. Since one frustum voxel may contain multiple 3D points, we need an efficient and effective way to aggregate features. Aliev ~\etal~\cite{aliev2019neural} propose to take the feature from the point closest to camera along the ray, however, this is not robust against geometry error and may result in temporal jitters or accidentally wrong occlusions.
In contrast, we maintain all the features available in the camera frustum thanks to the 3D volume.
To achieve sub-voxel performance, we vote each point feature to nearby voxels according to the distance to the voxel center.
Specifically, the feature in a voxel with coordinate in volume $(p,h,w)$ is calculated as:

\begin{equation}
    \label{equ: integration}
    F_{(p,h,w)}=\frac{\sum_iw^i_{(p,h,w)}\times F^i_{(p,h,w)}}{\sum_iw^i_{(p,h,w)}},
\end{equation}
\begin{equation}
    \label{equ: weight}
    w^i_{(p,h,w)}=(1-D^i_{1(p,h,w)})^a\times(\frac{1}{1+D^i_{2(p,h,w)}})^b,
\end{equation}
where $F^i$ is the feature of the $i$th point in a voxel, and $w^i_{(p,h,w)}$ is the blending weight of point $i$ to voxel $(p,h,w)$. $D_1\in(0,1/\sqrt{2})$ is the distance between the point projection on image to the corresponding pixel center of voxel $(p,h,w)$, and $D_2\in(0,l)$ is the depth difference between point and minimum depth point in this voxel.
Parameters $a$ and $b$ control the blending weight on direction parallel and vertical to the image plane. When $b\rightarrow+\infty$, it turns into Aliev \etal~\cite{aliev2019neural} where a point is picked via z-buffer. The motivation of Eq.~\ref{equ: weight} is to assign larger weights to points closer to camera or pixel center. We found it work well empirically. Other formulations reflecting similar property could also work properly.

\subsection{Multi-plane rendering}
We adopt the U-Net-like~\cite{ronneberger2015u} 3D convolutional neural network as the back-bone network. 
The 3D convolution effectively exploit information from neighboring pixel and depth, which naturally handle projection error caused by geometry noise.
In addition to that we also adopt dilated convolution in the last layer of encoder (left part of U-Net) to capture more image context. 
As the output of our network, we predict multi-plane RGB images plus their blending weights. The final output image $I$ is obtained by 
\begin{equation}
    \label{equ: weight_summation}
    I=\sum_p I_p\cdot\alpha_p,
\end{equation}
where $p$ indicates a plane, and $I_p$ and $\alpha_p$ are the corresponding plane predictions.
Please refer to supplementary material for more details of the network architecture.

\subsection{Loss function}
To measure the difference between network prediction and the ground truth image, we use perceptual loss~\cite{chen2018deep, johnson2016perceptual}, which we found practically works better than other common candidates, \eg $\ell_1$, $\ell_2$, SSIM.
In particular, we use feature vector from `input', `conv1-2', `conv2-2', `conv3-2', `conv4-2', `conv5-2' layers from a VGG-19 pre-trained on ImageNet dataset~\cite{deng2009imagenet}.
The perceptual loss is defined as a weighted sum of the $\ell_1$ loss on each feature maps. Specifically, the loss function is defined as follows: 
\begin{equation}
    \label{equ: loss}
    L(P_f,\theta)=\sum_l\lambda_l\|\Phi_l(I_g)-\Phi_l(f(P_f;\theta))\|_1,\vspace{-1.5ex}
\end{equation}
where $P_f$ represents point features, $I_g$ is ground-truth image, $\theta$ is network parameters, $f$ is our point cloud renderer, $\phi_l$ is a set of VGG-19 layers and $\lambda_l$ is weight used to balance different layers.

\subsection{Feature optimization.}
Inspired by Thies \etal~\cite{thies2019deferred} and Aliev \etal~\cite{aliev2019neural}, the appearance feature on each point can be updated via back-propagation.
Note that the aggregated voxel features are a weighted combination of point features, and thus gradient on the point feature using the chain rule is $-l_{r}\times w^i_{(p,h,w)}\times g_{(p,h,w)}$.
where $g_{(p,h,w)}$ is gradient derived from the loss function and $l_{r}$ indicates the learning rate. 
\section{Experiments}   
We evaluate our framework on various datasets and show qualitative and quantitative results. Particularly, we test the system robustness against noise in data that heavily degrades performance of previous methods.

\subsection{Datasets.} 
\begin{table*}[!htb]
    \centering
    \begin{tabular}{c|ccc|ccc}
    \hline
    Datasets & \multicolumn{3}{c}{ScanNet~\cite{dai2017scannet}} & \multicolumn{3}{c}{Matterport 3D~\cite{chang2017matterport3d}}\\
    \hline
    Methods & PSNR$\uparrow$ & SSIM$\uparrow$ &LPIPS$\downarrow$ & PSNR$\uparrow$ & SSIM$\uparrow$&LPIPS$\downarrow$\\
    \hline
    pix2pix~\cite{isola2017image} &19.247 &0.731 &0.429 &14.964 & 0.530 & 0.675\\
    Neural point based graphic~\cite{aliev2019neural} & \textbf{22.911}&\textbf{0.840}&0.245&17.931&0.622&0.597\\
    \hline
    Ours + direct render & 22.259&0.818&0.290&17.833&0.601&0.610\\
    \hdashline
    Ours &22.813 &0.835&\textbf{0.234}&\textbf{18.09}&\textbf{0.649}&\textbf{0.534}\\
    \hline
    \end{tabular}
    \caption{PSNR, SSIM and LPIPS values on ScanNet and Matterport 3D datasets.}
    \label{tab:SSIM PSNR}
\end{table*}

\paragraph{ScanNet~\cite{dai2017scannet}} contains RGBD scans of indoor environments. We follow the training and testing split of Aliev \etal~\cite{aliev2019neural}.
In particular, one frame is picked from every 100 frames for testing (e.g., frame 100, 200, 300...). The rest of frames are used for training. To avoid including frames that are too similar to the training set, the neighbors (with in 20 frames) of every testing frame are removed from training. 
Regarding the scene point cloud, we randomly lift 15\% pixels from the depth map into the 3D space to create a point cloud, which contains around 50 million points per scene. We then simplify it using volumetric sampling, leading to 8.9 million points per scene in average. 

\paragraph{Matterport 3D~\cite{chang2017matterport3d}} contains RGBD panoramas captured at multiple locations in indoor scenes. Each panorama is composed of 18 regular RGBD images viewing toward different directions. 
For each scene, we randomly pick 1/100 of the views for testing and leave the others for training.
Note that overall this dataset is more challenging due to the sparse point cloud and large variation of camera viewpoints. 

\subsection{Data preparation and training details}
For each scene, our network is trained for 21 epochs over 1,925 images on average, using Adam optimizer~\cite{kingma2014adam}. During training, $l_r$ is initialized as 0.01, which will be decreased every 7 epochs, and the learning rate decreasing follows $0.01\rightarrow0.005\rightarrow0.001$. 
$(P, H, W)$ are set to $(32, 480, 640)$ for ScanNet dataset and $(32, 512, 640)$ for Matterport 3D dataset according to the image resolution provided by the datasets. Point feature dimension is set as 11 and initialized as 0.5 (5 dimensions) + RGB (3 dimensions) + viewpoint direction (3 dimensions), note that only the first 8 dimensions will be updated. The hypeparameters $a, b$ in Eq.~\ref{equ: weight} are set to 1, and $\lambda_l$ in Eq.~\ref{equ: loss} following Chen \etal.~\cite{chen2018deep}. The training process is accomplished on one GeForce 1080 Ti, which takes on average 41.5 hours per scene. 

\begin{figure}[t]
 \centering
 \includegraphics[width=1.0\linewidth]{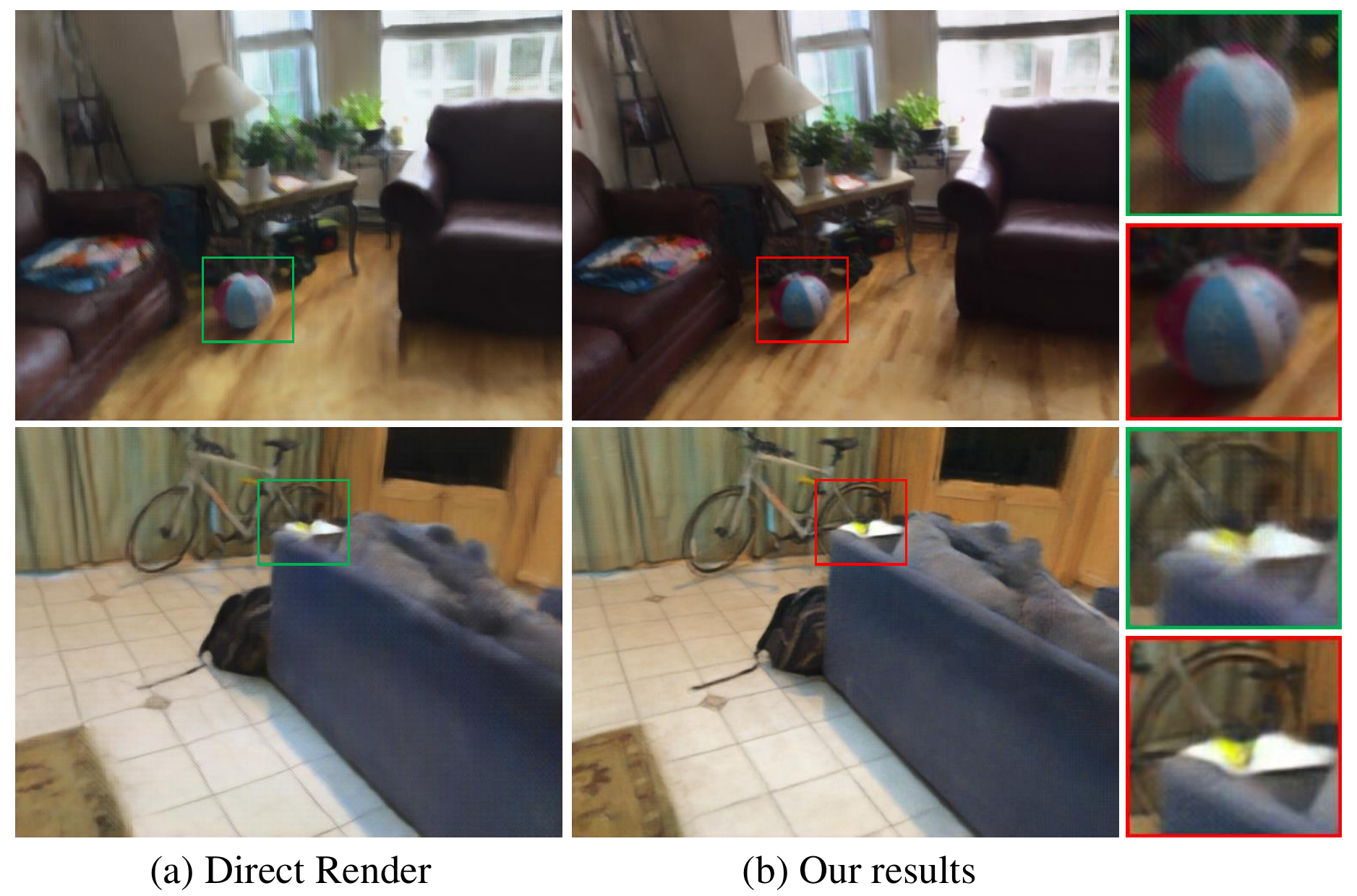}
 \caption{Advantage of neural descriptor. (a) displays results generated by directly rendering. (b) Ours results using neural descriptor. Without the assistant of neural descriptor, the final point clouds rendering results are blurry. Please zoom in for details.}
 \label{fig:descriptor}
 \vspace{-0.3cm}
\end{figure}

\subsection{Rendering results}

\begin{figure*}[!htb]
\centering
\includegraphics[width=0.9\linewidth]{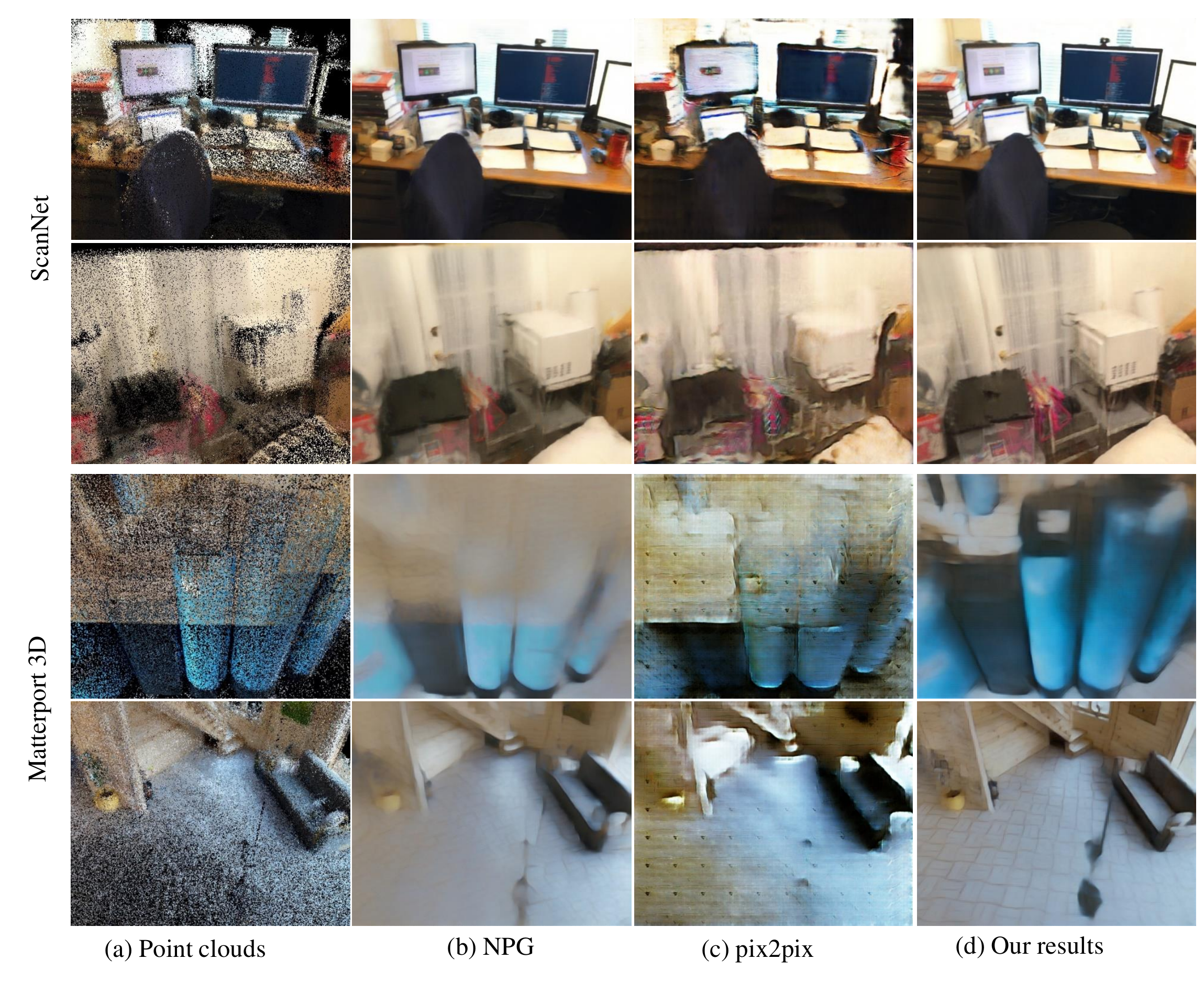}
\caption{Rendering quality on ScanNet and matterport 3D datasets. We compare our proposed method with pix2pix and neural point based graphic on two indoor scene datasets. By analyzing the generated image in novel view, our proposed method achieves better performance. Please zoom in for details.}
\label{fig:scannet_and_matterport}
\vspace{-0.3cm}
\end{figure*}

We first compare our method to two competitors, Neural Point-based Graphic(NPG)~\cite{aliev2019neural} and Pix2Pix~\cite{isola2017image}.
Specifically, NPG is a deep rendering approach that project 3D point features onto 2D image planes via a z-buffer and run 2D convolutions for neural rendering. They adopt U-Net like structure with gate convolution~\cite{yu2019free}.
Since the authors of~\cite{aliev2019neural} did not release the code, we implemented their method and achieved similar performance on the same testing cases. 
The Pix2Pix is an image to image translation framework~\cite{isola2017image}. 
The network takes projected colored point cloud and is trained to produce the ground truth. Compared to NPG, this baseline does not save a feature per point.

\subsubsection{Quantitative comparisons}
We use standard metric, Peak Signal-to-Noise Ratio (PSNR) and Structural Similarity Index (SSIM), to measure the rendering quality.
Since these two metrics may not necessarily reflect the visual quality, we also adopt a human perception metric, Learned Perceptual Image Patch Similarity (LPIPS)~\cite{zhang2018unreasonable}. 
Table~\ref{tab:SSIM PSNR} reports the comparison on two datasets. 
Our method is significantly better than Pix2Pix on both dataset with a large improvement margin on all the metrics.
When compared to NPG, our method outperforms on Matterport3D, and is comparable on ScanNet, where NPG achieves better PSNR and SSIM, while our LPIPS is higher.
Some examples from both dataset are shown in Fig~\ref{fig:scannet_and_matterport}.    
Note that it has been mentioned in their own paper that NPG is optimized for pixel-wise color accuracy at the cost of sacrificing the temporal consistency. In contrast, our result is free such jittering, especially obvious at depth boundaries. 
Please refer to the supplementary video for visual comparisons.


\noindent
\paragraph{Direct render}
To verify if learning point cloud feature is necessary. We train our model taking only the point RGB value as the feature, which is referred to as `direct render'. The results are displayed in Fig.~\ref{fig:descriptor}. 
As seen, the direct render without point feature is more blurry (\eg, the sofa) and lack of specular components (\eg, the ball). 
This indicates that point feature helps to encode material related information and support view-dependent components. We also report the quantitative numbers in Table~\ref{tab:SSIM PSNR} as `Ours+direct render'. It is observed that ours with learnt features outperforms the direct render approach on all metrics.

\subsubsection{Qualitative comparisons}
Figure~\ref{fig:scannet_and_matterport} shows some visual comparisons of our method with NPG and Pix2Pix on ScanNet and Matterport 3D datasets. The first two rows show two scenes of ScanNet while the third and forth rows show two scenes of Matterport 3D. Fig.~\ref{fig:scannet_and_matterport} (a) shows point cloud. The point cloud is noisy and incomplete. Fig~\ref{fig:scannet_and_matterport} (b) shows the result of NPG. For its ScanNet results, we notice some incorrect places, e.g., black stripe on the labtop screen of first scene, and missing details of shelf of second scene (Please zoom in for details). For its Matterport results, the missing of details become more serious, e.g. the floor textures is missing in the forth example. Fig~\ref{fig:scannet_and_matterport} (c) shows the result of pix2pix. It generates strange curves for ScanNet while introduce weird textures for Matteport result. Fig~\ref{fig:scannet_and_matterport} (d) shows our results. As seen, our result is free from such problems. 

\subsection{Robustness and stability}
\begin{figure*}[t]
\centering
\includegraphics[width=0.95\linewidth]{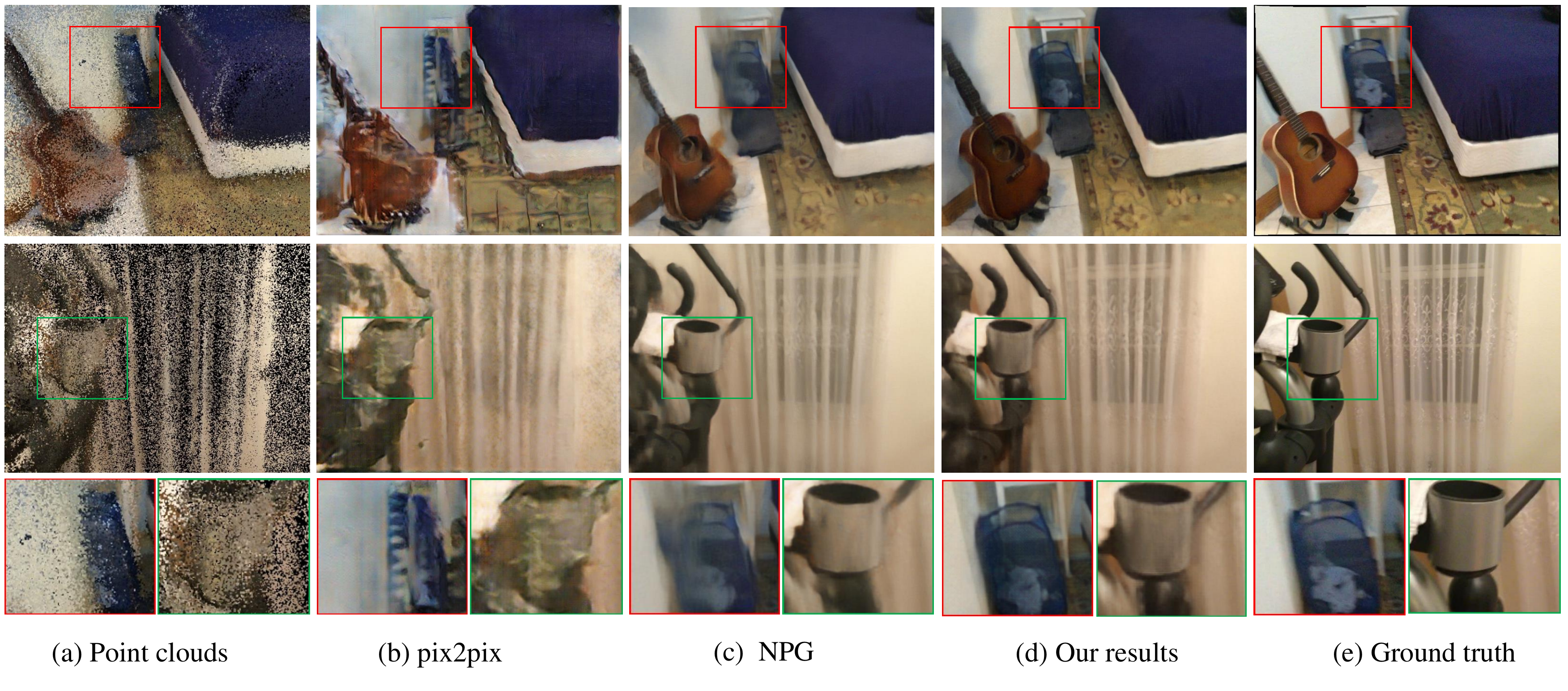}
\caption{Restore adversely occluded objects. (a) Noisy point clouds with objects adversely occluded. (b) pix2pix method. (c) Results generated through neural point based graphic method~\cite{aliev2019neural} from 2D point clouds images. Can't efficiently restore occluded objects. (d) Our results generated from 3D neural render. Restore adversely occluded objects. (e) Ground truth.}
\label{fig:occluded}
\end{figure*}

\begin{figure*}[t]
\centering
\includegraphics[width=0.95\linewidth]{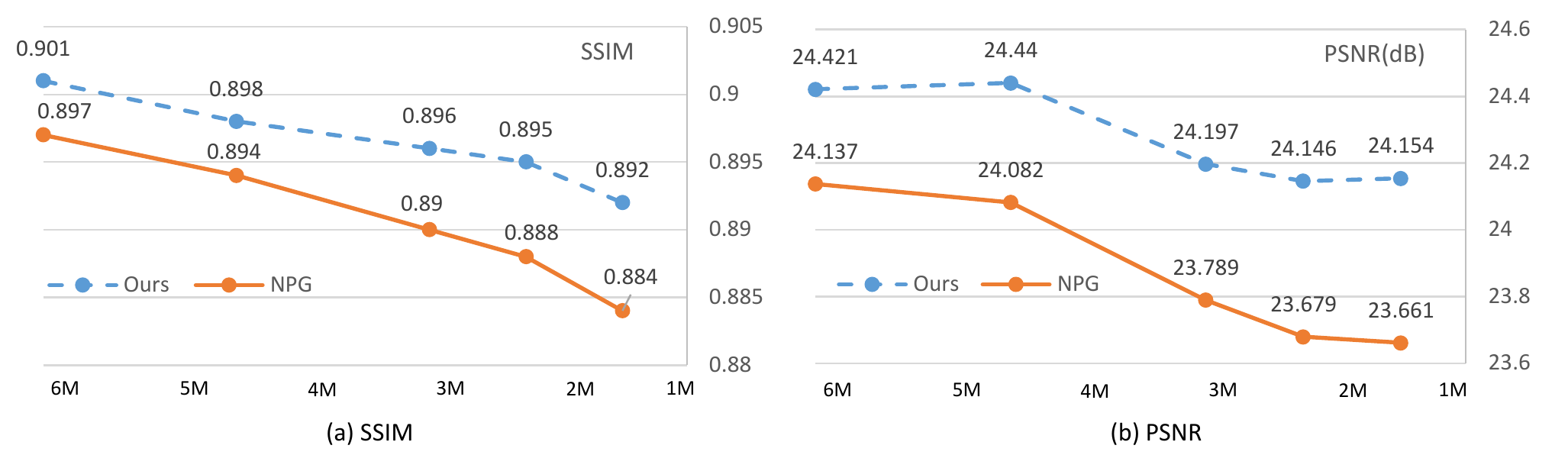}
\caption{SSIM and PSNR trend when point clouds density decreased.}
\label{fig:trend}
\vspace{-0.5cm}
\end{figure*}

\paragraph{Robustness to noise point cloud}
In practice, the point cloud are usually noisy, and the rendering model needs to  tolerant such noises to produce robust results.
When objects are close to each other, noisy depth may result in wrong z-buffer such that the correct points are occluded.
This is especially harmful for methods that rely on 2D projection of the point cloud, such as NPG and Pix2Pix.
In contrast, our method maintains all related point features in the camera frustum volume and allows network to infer correctly.
Fig.~\ref{fig:occluded} shows two comparisons on cases with noisy depth.
NPG and Pix2Pix either completely miss the correct objects or produce a mixture of foreground and background.

\begin{figure*}[t]
\centering
\includegraphics[width=.87\linewidth]{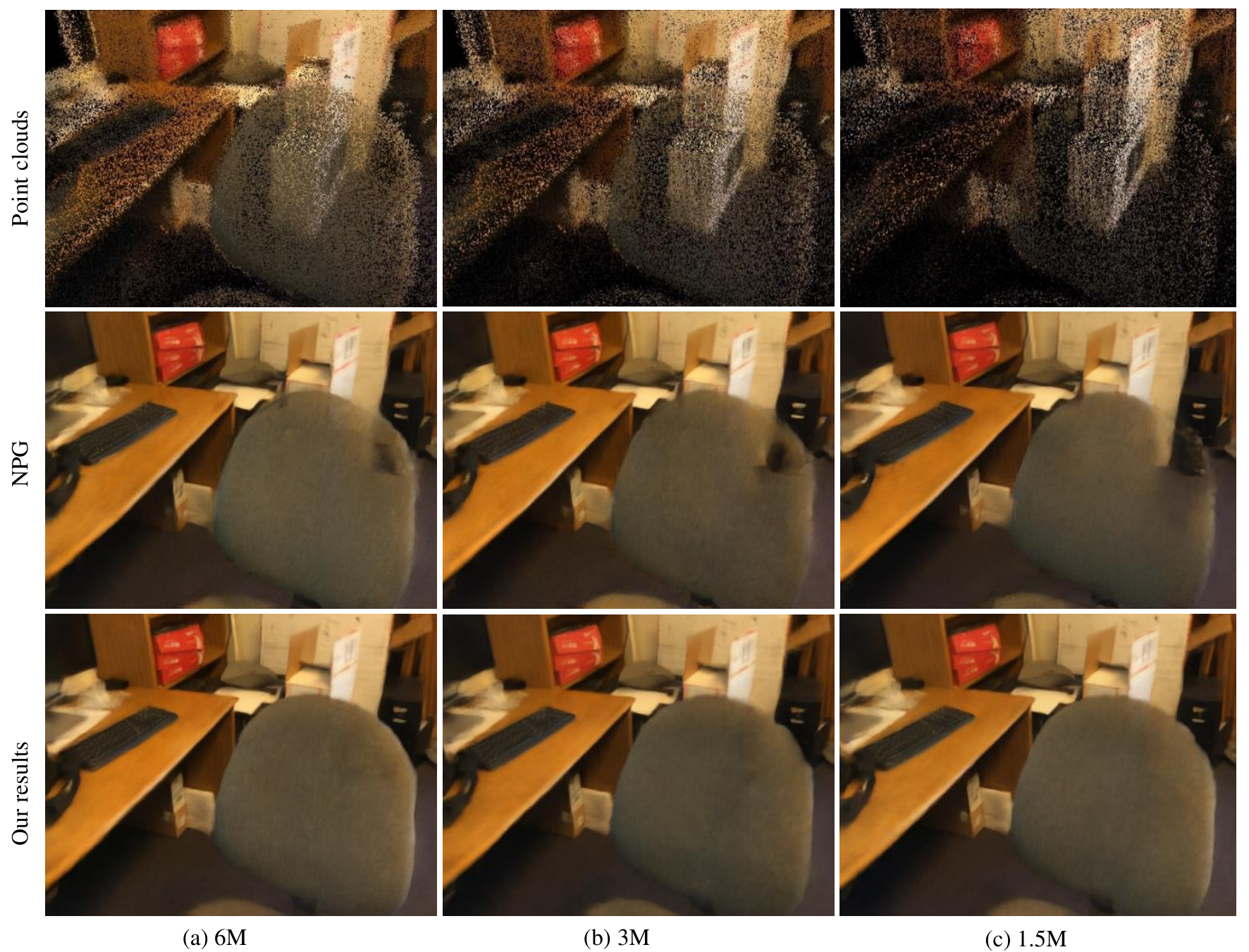}
\caption{Different sparsity of point clouds. The number of point clouds varies from column to column, and different row shows results generated from different method. By analysis, our method can work at relatively sparse point clouds.  }
\label{fig:sparsity}
\vspace{-0.4cm}
\end{figure*}

\paragraph{Robustness to varying densities}
Theoretically, our method can support arbitrarily large scene since the point features can be stored on hard drive.
However for efficient rendering, it is more desirable to keep point features in memory and render from relatively sparse point cloud to save both the memory and the computational cost for point projection.
Unfortunately, sparse point cloud may result in in-complete z-buffer such that the occluded background shows up in the image.
NPG proposes to assign each point a square size during the projection according to its depth to mitigate this issue, but may not be sufficient.
Fig.~\ref{fig:sparsity} shows a qualitative comparison to NPG on the same scene with different point density.
With less points, NPG reveals more background due to imcomplete z-buffer, while our method still maintain the chair.
Fig.~\ref{fig:trend} further shows the quantitative comparison.
As seen, while both methods perform worse with fewer points, the metrics of our method drops relatives slower, which means using camera frustum is more robust against the varying point density.

\paragraph{Temporal consistency}
3D camera frustum also helps to improve the temporal consistency. 2D projection based methods may project very different points for the same 3D location to very close camera viewpoints.
This is because the order of the points in the z-buffer may dramatically change between slightly different camera views.
Consequently, the rendering of the same 3D location may use features from different points and thus cause jittering artifacts.

\begin{figure}[t]
\centering
\includegraphics[width=1.0\linewidth]{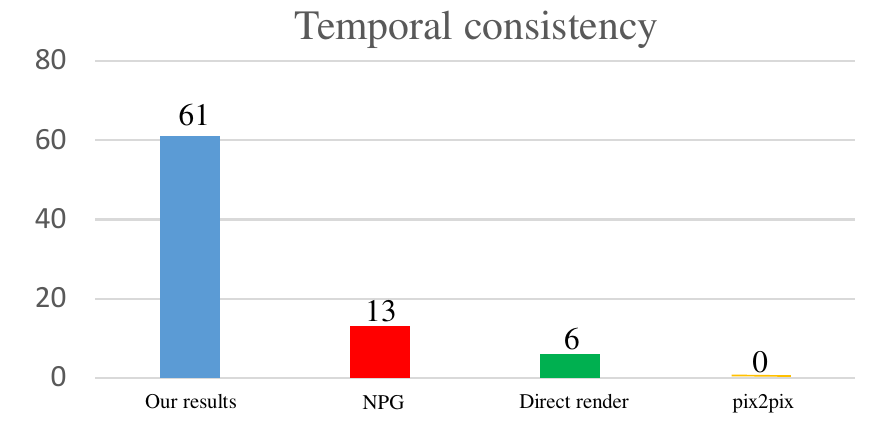}
\caption{The user study result on preference for videos synthesized by different methods (i.e. ours, NPG, direct render and pix2pix.)}
\label{fig: user_stdy}
\vspace{-0.4cm}
\end{figure}

We perform a user study to compare the temporal consistency against NPG , Pix2Pix and direct render. We render 4 videos of 4 different scenes with respect to each method. During user study, each time, we present 4 videos of 4 approaches and ask the subject to pick the best one. As we have 4 scenes, a user will pick 4 times. 20 users are invited, accumulating to 80 picks in total. The participants are required to only judge the temporal consistency. Statistical results are shown in Fig.~\ref{fig: user_stdy}. Our method received 61 picks, which indicates our video is apparently better than other methods in terms of temporal consistency. Please refer to the supplementary files for these videos.  

\section{Conclusion}
In this work, we have proposed a method which synthesizes novel view images from 3D point clouds. Instead of directly project features from 3D points onto 2D image domain, we projected these features into a layered volume of
camera frustum, such that the visibility of 3D points can be naturally maintained. Through experiments, our method is robust to point clouds noise and generates flicker-less videos. 
In the future, we will explore novel view synthesis from point clouds in multiple views. Optical flow can be utilized for enforcing temporal consistency given additional observations. Flickers can be removed by enforcing constraints on similar predictions from shared points. In addition, applying interpolation in different depth planes could further improve the robustness against sparse point clouds. 

\noindent\textbf{Acknowledgement: }This research was supported in part by National Natural Science Foundation of China (NSFC, No.61872067, No.61720106004), in part by Research Programs of Science and Technology in Sichuan Province (No.2019YFH0016).


\newpage


\section*{Supplementary material}

# Supplementary Materials for
# "Neural Point Cloud Rendering via Multi-Plane Projection"

Peng Dai[1*]    Yinda Zhang[2*]    Zhuwen Li[3*]    Shuaicheng Liu[1†]    Bing Zeng[1]

[1]University of Electronic Science and Technology of China
[2]Google Research    [3]Nuro Inc

We provide details about network architecture, evaluation details, other comparisons and results on various datasets.

## 1. Network Architecture

The network architecture is provided in Table. 1, including kernel size, stride, kernel dilation, activation function, and feature channels. In short, our network is a UNet with short-cut connection.

## 2. Evaluation Details

Our test set consists of four scenes from ScanNet [3] and two scenes from Matterport3D [2]. The scene IDs chosen from ScanNet are 'scene0000_00', 'scene0010_00', 'scene0016_00', 'scene0024_00', and the first and fourth scenes are also adopted in NPG [1]. The scene IDs chosen from Matterport3D are '17DRP5sb8fy' and '29hnd4uzFmX'.

We calculate PSNR and SSIM in standard way by using the function provided by Matlab.

## 3. More Results

### 3.1. Multi-Plane Image and Weight

For the network outputs, we produce multiple planes of images with their blending weights. Fig 1 visualizes one of the output example. As can be seen, the blending weights roughly reflects the depth of the scene since each plane corresponds to a specific depth in camera coordinate. Regarding the color image, all the images seems to maintain the scene layout reasonably well, while for each pixel the plane with the correct depth are likely to provide the right color.

### 3.2. Compare with image inpainting

An alternative approach for point cloud rendering is to

---

*Equal contribution
†Corresponding author

---

treat it as an image inpainting problem, which renders an incomplete 2D image from the sparse point cloud and fill the missing region. To compare with this baseline, we adopt U-Net as backbone, and replace ordinary convolution operation with gate convolution (GateConv) [6] and sparse convolution (SparseConv) [5], which were proposed in SOTA image inpainting methods to fill up irregular holes. We compare our method to this image inpainting baseline, and the results are shown in Fig 2. We find that directly inpaint sparse point clouds images (Fig 2(a)) generates blurry results (Fig 2(b)(c)) compared to our results (Fig 2(d)). This is presumably because that the color of directly projected points are different with ground truth images due to view-dependent effects (e.g. specular) and different exposure time.

### 3.3. Textured mesh refinement

Even though choosing point cloud as intermediate is flexible and saves time from mesh reconstruction, we also compare to a mesh rendering based approach. We first render the textured mesh of the scene into the target camera viewpoint (Fig 3(a)) and utilize neural network to refine them. The comparison is shown in Fig 3(b). After refinement, the missing region in textured mesh (re-projected) can be roughly completed, but has less details than our results (Fig 3(c)). Moreover, the PSNR and SSIM of refined images (re-projected textured mesh) are 22.208 and 0.86 respectively in 'scene0010_00', and our results are 24.421 and 0.901.

### 3.4. Test on Creepy Attic

We test our method on Creepy Attic [4] which consists of kinect captured RGB-D and digital camera captured high quality RGB images. In Hedman et al. [4], camera parameters are generated via structure from motion (SFM), and per-view depth images are obtained using multi-view stereo (MVS). Due to the limited number of high-quality images, our method firstly train on kinect captured images, then fine-tune on digital camera captured RGB images. Fig 4 show some test results (randomly selected poses that never

| Layers | kernel size | stride | dilation | channel_in | channel_out | actv | inputs |
|---|---|---|---|---|---|---|---|
| conv0 | 1×1×1 | 1×1×1 | 1 | 11 | 16 | lrelu | neural point features |
| conv1 | 3×3×3 | 1×1×1 | 1 | 16 | 32 | lrelu | conv0 |
| maxpool1 | 2×2×2 | 2×2×2 | - | 32 | 32 | - | conv1 |
| conv2 | 3×3×3 | 1×1×1 | 1 | 32 | 32 | lrelu | maxpool1 |
| maxpool2 | 2×1×1 | 2×1×1 | - | 32 | 32 | - | conv2 |
| conv3 | 3×3×3 | 1×1×1 | 1 | 32 | 64 | lrelu | maxpool2 |
| maxpool3 | 2×2×2 | 2×2×2 | - | 64 | 64 | - | conv3 |
| conv4 | 3×3×3 | 1×1×1 | 1 | 64 | 64 | lrelu | maxpool3 |
| maxpool4 | 2×1×1 | 2×1×1 | - | 64 | 64 | - | conv4 |
| conv5 | 3×3×3 | 1×1×1 | 1 | 64 | 128 | lrelu | maxpool4 |
| maxpool5 | 2×2×2 | 2×2×2 | - | 128 | 128 | - | conv5 |
| conv6 | 1×3×3 | 1×1×1 | 2 | 128 | 128 | lrelu | maxpool5 |
| up1 | 2×2×2 | 2×2×2 | 1 | 128 | 128 | - | conv6 |
| conv7 | 3×3×3 | 1×1×1 | 1 | 256 | 128 | lrelu | up1+conv5 |
| up2 | 2×1×1 | 2×1×1 | 1 | 128 | 64 | - | conv7 |
| conv8 | 3×3×3 | 1×1×1 | 1 | 128 | 64 | lrelu | up2+conv4 |
| up3 | 2×2×2 | 2×2×2 | 1 | 64 | 64 | - | conv8 |
| conv9 | 3×3×3 | 1×1×1 | 1 | 128 | 64 | lrelu | up3+conv3 |
| up4 | 2×1×1 | 2×1×1 | 1 | 64 | 32 | - | conv9 |
| conv10 | 3×3×3 | 1×1×1 | 1 | 64 | 32 | lrelu | up4+conv2 |
| up5 | 2×2×2 | 2×2×2 | 1 | 32 | 32 | - | conv10 |
| conv11 | 3×3×3 | 1×1×1 | 1 | 64 | 32 | lrelu | up5+conv1 |
| rgbs | 1×1×1 | 1×1×1 | 1 | 32 | 3 | - | conv11 |
| blend weights | 1×1×1 | 1×1×1 | 1 | 32 | 1 | softmax | conv11 |

Table 1. Details of network architecture, where **actv** is the activation function. And **up1-5** are transposed convolutions.

### 3.5. More Results on ScanNet

In this section, we show more results on ScanNet dataset in Fig 5 and Fig 6. Our method is better than pix2pix in terms of the visual quality. Compared to NPG [1], we perform especially better at the location with sparsity and occlusion.

### 3.6. More Results on Matterport3D

We also show more results on Matterport3D dataset in Fig 7 and Fig 8. Our results are significantly better than all the other methods. Our camera frustum based point projection is more robust compared to z-buffer based projection [1] when the point cloud is sparse, which is especially true on large scenes from Matterport3D. As a result, our renders are more complete and reflect correct occlusion.

### 3.7. Video for Temporal Consistency

To demonstrate the temporal coherence, we provide video sequences on the same trajectories from NPG and our method. We compare video temporal consistency synthesized by different methods (e.g. ours, pix2pix, neural point based graphic, direct render.) in different scenes.

appear during training).

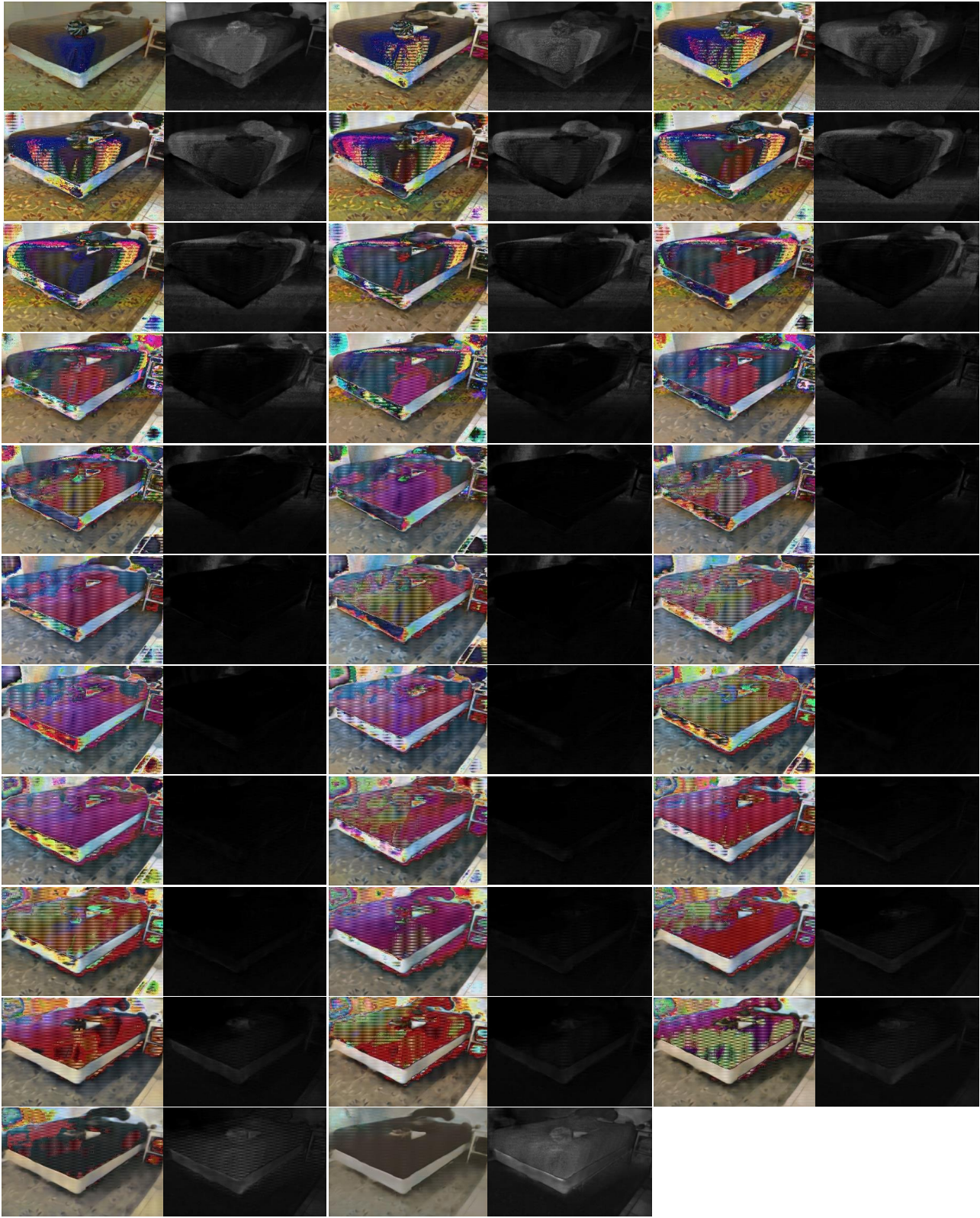

Figure 1. One example of multi-plane RGB images and blending weights. Left to right, up to down.

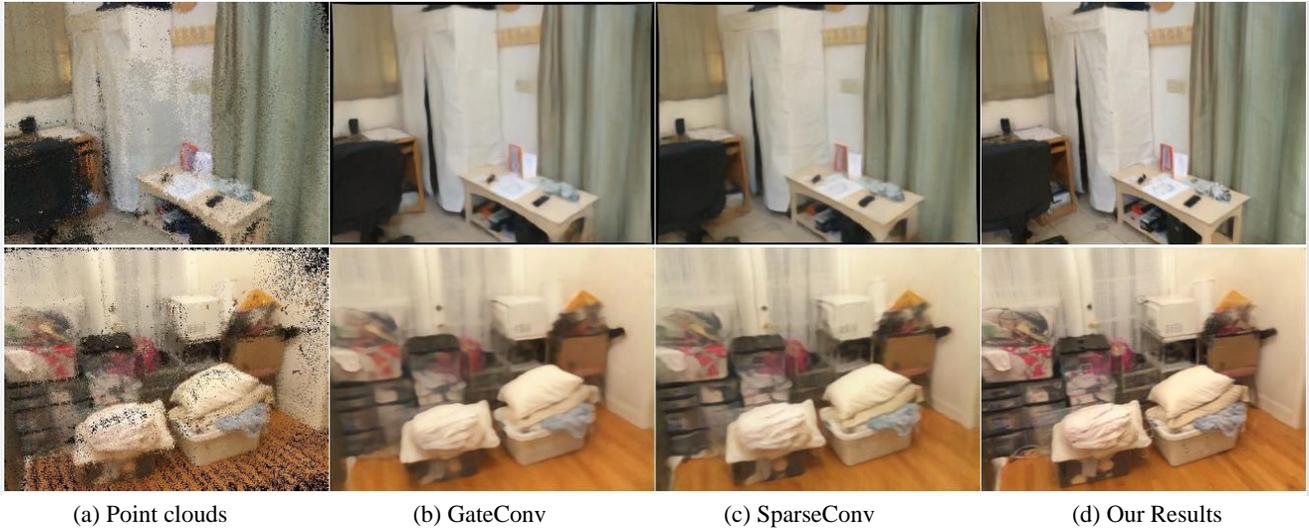

(a) Point clouds  (b) GateConv  (c) SparseConv  (d) Our Results

Figure 2. Compare with image inpainting. (a) Re-projected 2D point clouds images with holes. (b) Results generated by using gate convolution. (c) Results generated by using sparse convolution. (d) Results generated by our method. Directly apply image inpainting will generate blurry results. (Zoom in for details)

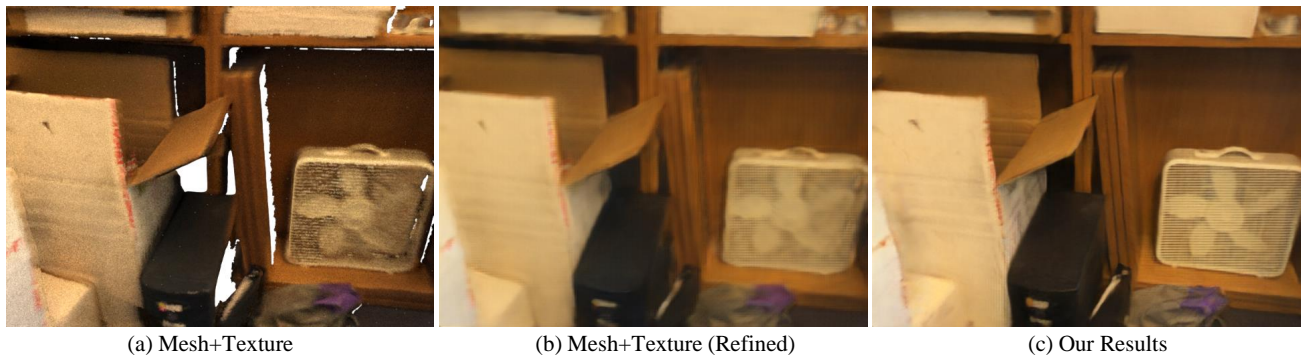

(a) Mesh+Texture  (b) Mesh+Texture (Refined)  (c) Our Results

Figure 3. Compare with textured mesh refinement. (a) Re-project textured mesh into 2D image plane. (b) Refined result through neural network. (c) Result generated by our method. The missing region in (a) can be roughly completed through network, but the refined image (b) has less details compared with our results (c). (Zoom in for detail)

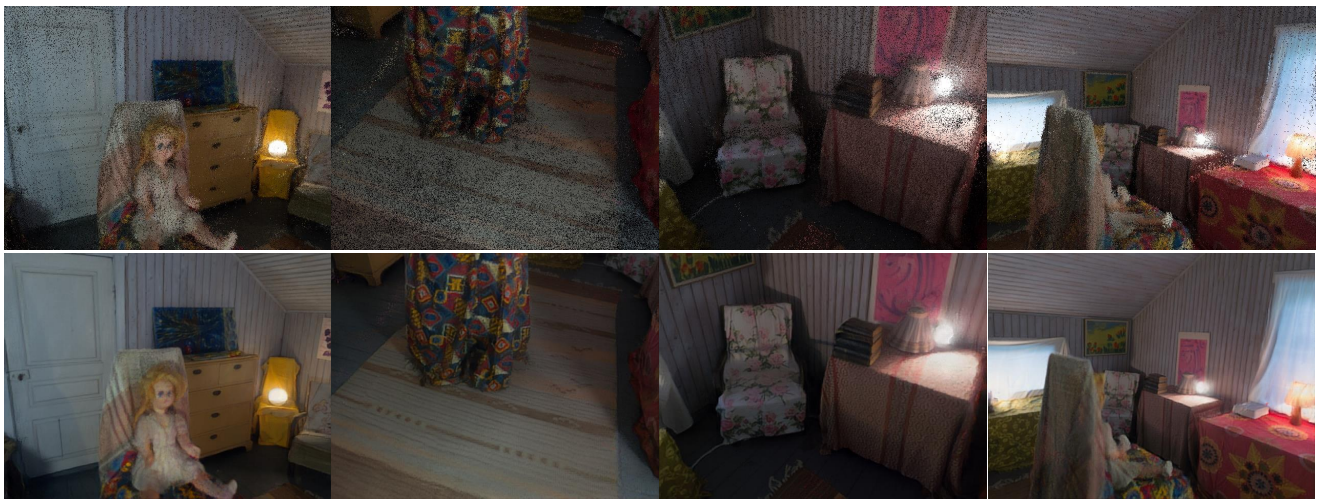

Figure 4. Test results(second row) on Creepy Attic with corresponding colored point clouds(first row).

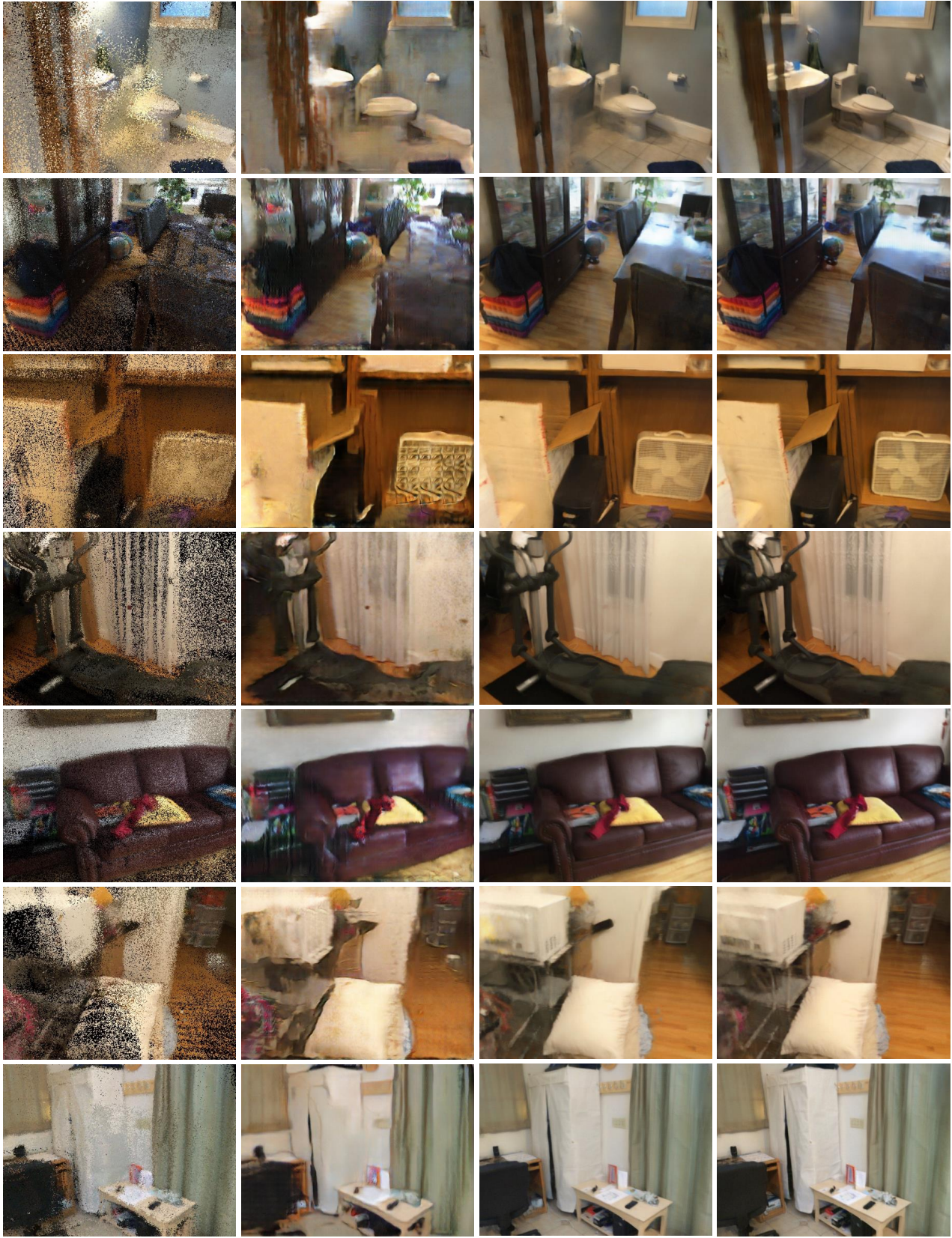

(a) Point clouds    (b) pix2pix    (c) NPG    (d) Our results

Figure 5. More results on ScanNet. Our results are better than other methods, especially at the location with sparsity and occlusion. Zoom in for details.

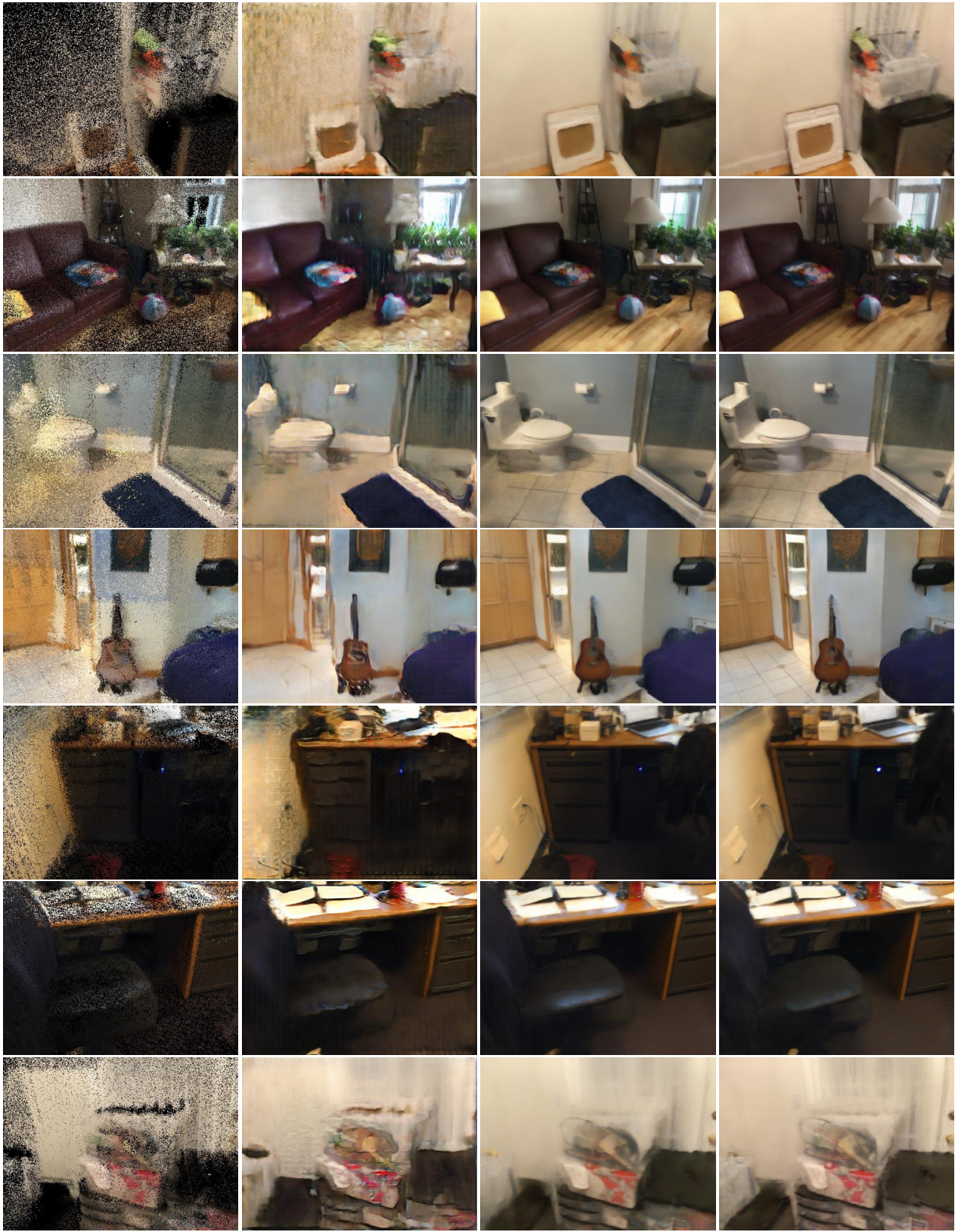

(a) Point clouds　　　　(b) pix2pix　　　　(c) NPG　　　　(d) Our results

Figure 6. More results on ScanNet. Our results are better than other methods, especially at the location with sparsity and occlusion. Zoom in for details.

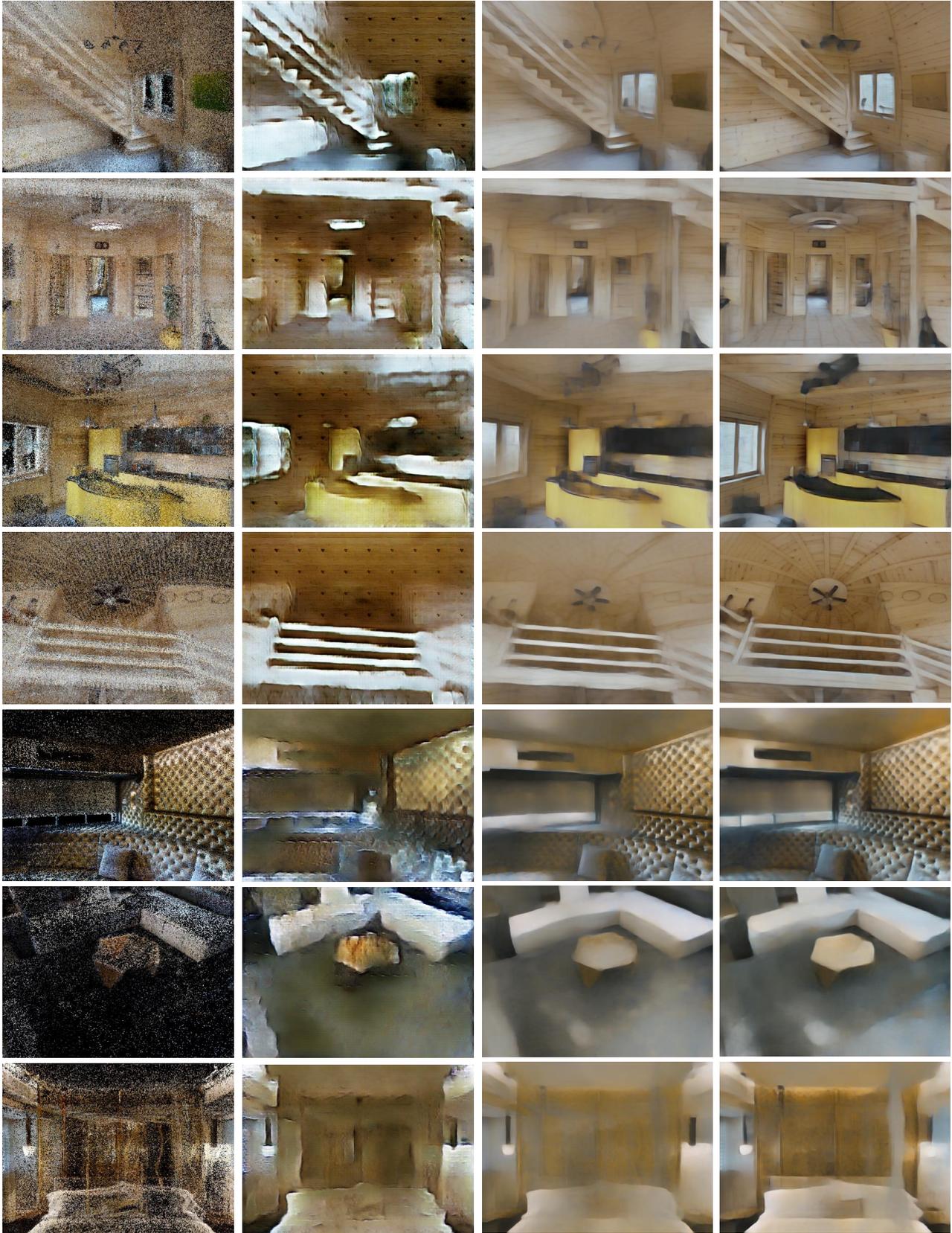

(a) Point clouds  (b) pix2pix  (c) NPG  (d) Our results

Figure 7. More results on Matterport3D. Our method obtains higher performance than other baselines on this challenging dataset.

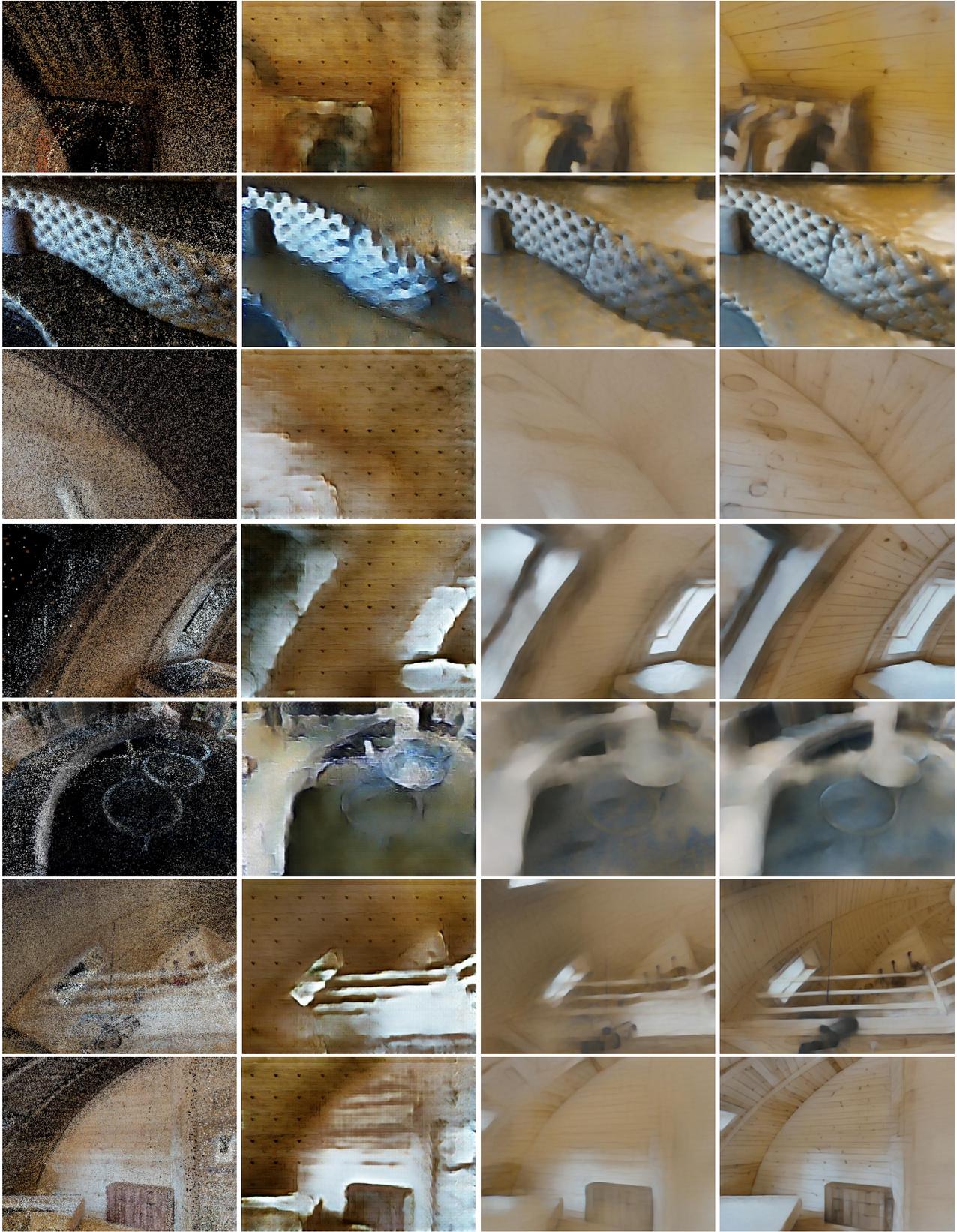

(a) Point clouds　　(b) pix2pix　　(c) NPG　　(d) Our results

Figure 8. More results on Matterport3D. Our method obtains higher performance than other baselines on this challenging dataset.


\end{document}